\documentclass[12pt]{spieman}  
\usepackage{amsmath,amsfonts,amssymb}
\usepackage{graphicx}
\usepackage{setspace}
\usepackage{tocloft}
\usepackage{multirow}

\title{City Scene Super-Resolution via Geometric Error Minimization}

\author[a]{Zhengyang Lu}
\author[a,*]{Feng Wang}
\affil[a]{Jiangnan University, School of Design, Wuxi, China}

\cftpagenumbersoff{figure}
\cftpagenumbersoff{table} 
\begin{document} 
\maketitle

\begin{abstract}
Super-resolution techniques are crucial in improving image granularity, particularly in complex urban scenes, where preserving geometric structures is vital for data-informed cultural heritage applications. In this paper, we propose a city scene super-resolution method via geometric error minimization. The geometric-consistent mechanism leverages the Hough Transform to extract regular geometric features in city scenes, enabling the computation of geometric errors between low-resolution and high-resolution images. By minimizing mixed mean square error and geometric align error during the super-resolution process, the proposed method efficiently restores details and geometric regularities. Extensive validations on the SET14, BSD300, Cityscapes and GSV-Cities datasets demonstrate that the proposed method outperforms existing state-of-the-art methods, especially in urban scenes. 
\end{abstract}


{\noindent \footnotesize\textbf{*}Feng Wang,  \linkable{wangfeng@jiangnan.edu.cn} }

\begin{spacing}{2}   

\section{Introduction}
With the rapid development of cultural heritage preservation and urban planning \cite{hou2022digitizing,galani2019evaluating,hug2012qualitative}, super-resolution techniques have attracted significant attention, particularly for enhancing the granularity of city scene imagery\cite{chen2022real}. 
Low-resolution city images obscure valuable details, challenging the recognition and interpretation of architectural and cultural elements.
This limitation affects various applications, such as heritage monitoring, virtual tourism, and cultural preservation. 

Single image super-resolution (SISR) is a crucial approach in computer vision to enhance image resolution  \cite{zhou2020single,song2021gradual,lu2023joint}.
However, the reconstruction of high-resolution images from low-resolution inputs remains challenging due to the inherent ambiguity caused by multiple possible super-resolution mappings \cite{shang2016understanding,lu2022pyramid}.
Existing SISR methods, including internal and external learning-based approaches, have demonstrated effectiveness in certain scenarios but may lack the ability to accurately preserve intricate geometric structures found in city scenes.

\begin{figure}[htbp]
	\centering
	\includegraphics[width=\linewidth]{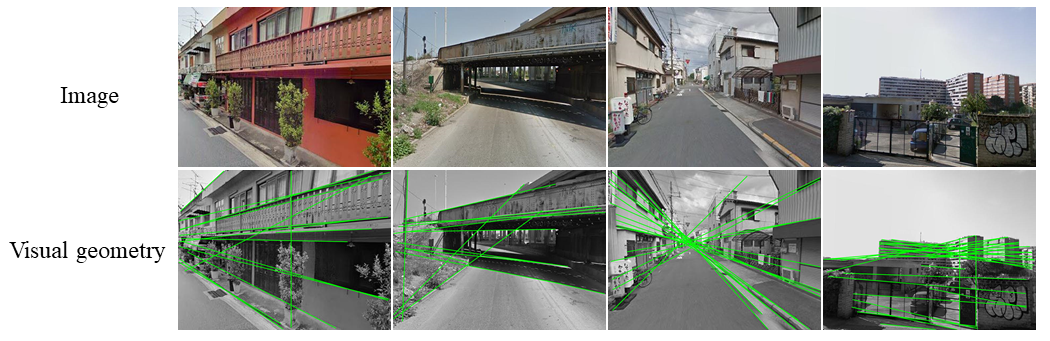}
	\caption{Visual representations of urban scenes highlighting the geometric features resulting from Hough transform.}
	\label{fig:geodemo} 
\end{figure}

To address the limitations of conventional SISR methods in city scenes, this paper proposes a novel single-image super-resolution method designed specifically for urban environments. Our approach utilizes the Hough Transform, a linear pattern recognition algorithm, to extract regular geometric objects and lines commonly found in city landscapes. The rich geometric regularities in urban scenes, such as buildings and traffic objects, provide essential context for guiding super-resolution algorithms. Figure \ref{fig:geodemo} shows visual representations of city scenes and geometric features.

The proposed method aims to minimize the geometric error between low-resolution and high-resolution images during the super-resolution process. By prioritizing geometric accuracy alongside pixel-level details, we aim to achieve higher-resolution images that maintain the geometric structures, thus enabling accurate cultural heritage representations.

This paper makes three key contributions:

\begin{itemize}
	\item A novel method that minimizes geometric errors in the super-resolution process, significantly improving accuracy in the representation of cultural heritage in high-resolution images;
	
	\item The Hough transform, utilized for geometric feature extraction, is applied to super-resolution tasks, providing geometric constraints to the neural networks in city scenes.
	
	\item Extensive validation on the Cityscapes\cite{cordts2016cityscapes} and Google Street View datasets\cite{ali2022gsv}, showcasing the superior performance of the proposed approach compared to state-of-the-art methods.
\end{itemize}

The paper is organized as follows: Section II provides an overview of related works in super-resolution; Section III analyses the existing problems in the super-resolution field. Section IV details the proposed method, including network structure and geometric constraint; Section V presents the experimental results and comparisons with existing methods; and finally, Section VI concludes the paper and discusses potential avenues for future research.

\section{Related Works}

Super-resolution techniques aim to infer the high-resolution image from the low-resolution image~\cite{irani2009super}. Deep-learning methods brought a paradigm shift in single-image super-resolution techniques. The section below concisely summarises various deep-learning SISR methodologies that have significantly influenced the field.

Numerous super-resolution techniques can be classified into two categories: video super-resolution reconstruction, exemplified by methods like VESPCN~\cite{caballero2017real}, and SISR, which comprises widely-used methods including SRCNN~\cite{dong2015image}, FSRCNN~\cite{dong2016accelerating}, VDSR~\cite{kim2016accurate}, CARN~\cite{ahn2018fast}, DRCN~\cite{kim2016deeply}, SRGAN~\cite{ledig2017photo}, ESPCN~\cite{shi2016real}, EDSR~\cite{lim2017enhanced}, NLSAN~\cite{mei2021image}, and DBPN~\cite{haris2018deep}.

He~\cite{dong2015image} initially proposed SRCNN, pioneering the use of deep learning to solve super-resolution problems. Despite the groundbreaking approach, SRCNN faced significant challenges, namely a restricted field of view and a tendency towards over-fitting, rendering the training process arduous. To overcome these limitations, FSRCNN~\cite{dong2016accelerating} was introduced, implementing three fundamental modifications compared to SRCNN. It adopted the original low-resolution image as input, incorporated a deconvolution layer for upsampling at the network's end, and utilized smaller filter kernels with a deeper network architecture for the super-resolution task.

To improve the capacity of neural networks, Kim~\cite{kim2016accurate} proposed VDSR, a deeper network that employed residual learning and gradient clipping to counteract the slow convergence issues typically associated with a large number of parameters. In addition, DRCN~\cite{kim2016deeply} developed an inference network for non-linear feature mapping, where an interpolated image was used as input, and a recursive network structure was implemented for data processing.

SRGAN~\cite{ledig2017photo}, the first Generative Adversarial Network (GAN) designed for super-resolution tasks, incorporated a deep residual network with skip-connections and introduced a perceptual loss function, thereby improving the visual quality of super-resolution reconstructions.
ESPCN~\cite{shi2016real} was designed to combat the significant computational complexity of deep networks. It introduced a sub-pixel convolutional layer, which notably enhanced the efficiency of the deconvolution operation, thereby boosting the overall performance of SISR tasks.
EDSR~\cite{lim2017enhanced}, winner of the NTIRE-2017 Super-Resolution Challenge Competition, improved performance significantly by removing batch normalization layers from SRResNet. This change allowed for the model to be scaled up, thereby enhancing the quality of the results.
Haris~\cite{haris2018deep} introduced DBPN, which employed iterative upsampling and downsampling and implemented an error feedback mechanism. It built interconnected upsampling and downsampling blocks representing different aspects of the low-resolution and high-resolution components.
CARN~\cite{ahn2018fast} utilized a cascading mechanism at both local and global levels to consolidate features from multiple layers. This design allowed for a more comprehensive range of input representations to be captured, thereby enhancing the model's performance.

To construct object sharp edges, Lu proposed UnetSR\cite{lu2022single} with the mixed gradient error that blends mean squared error and mean gradient error. Furthermore, Lu introduced a more scalable technique for feature extraction from shallow layers, employing the shuffle pooling method in dense U-Net model\cite{lu2022dense}.
Recent progress includes FKP~\cite{liang2021flow}, a super-resolution kernel modelling technique based on normalizing flow, and NLSAN~\cite{mei2021image}, a method that offers a dynamic sparse attention pattern. FKP enhances the kernel in the latent space, while NLSAN improves non-local attention through spherical locality-sensitive hashing, resulting in a segmentation of the input space into associated feature hash buckets.

\section{Problem Analysis}

\begin{figure}[htbp]
	\centering
	\includegraphics[width=\linewidth]{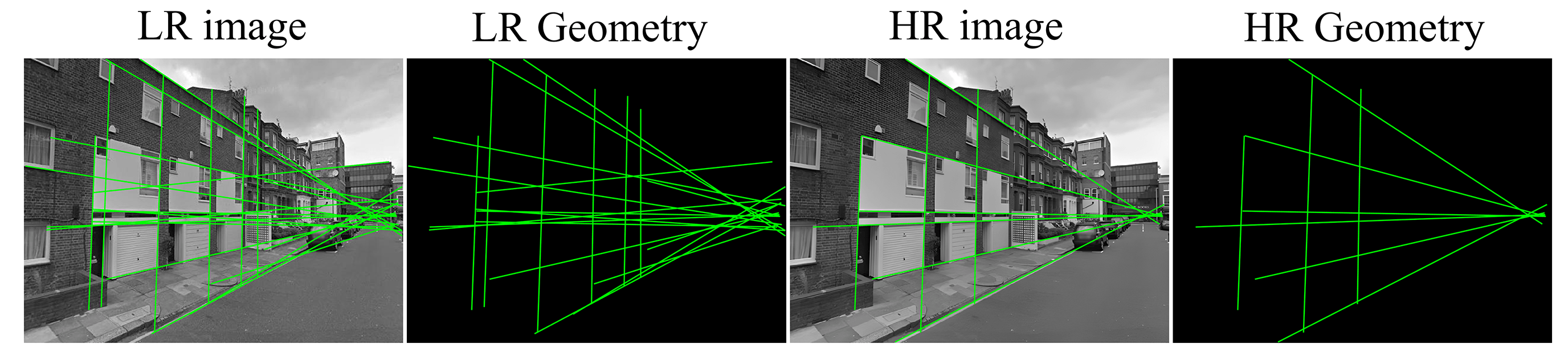}
	\caption{Visual results and geometric features for high-resolution images and SRGAN model reconstructed super-resolution images, with geometric features extraction by Hough transform.}
	\label{fig:problem} 
\end{figure}

Existing super-resolution models, while powerful, suffer from two primary shortcomings. 
First, previous works focus on pixel relations rather than inherent geometric form. 
The pixel-centric approach often results in high-resolution images that, while detailed, demonstrate significant discrepancies in geometric consistency when compared to low-resolution originals, as shown in Figure \ref{fig:problem}. 
The inherent geometric regularities of artificial buildings compound the challenge, necessitating an practical approach that values geometric forms in super-resolution.

Second, most existing super-resolution models adopt a one-size-fits-all approach, rendering models generic in nature. Although most generic models can handle various real-world scenes, models' performance is often weakened in specific scenarios. For example, when it comes to city scene imagery crucial to cultural heritage applications, most models fail to capture unique elements such as complex architectural styles or irregular urban landscapes.

\section{Methodology}

\begin{figure}[htbp]
	\centering
	\includegraphics[width=\linewidth]{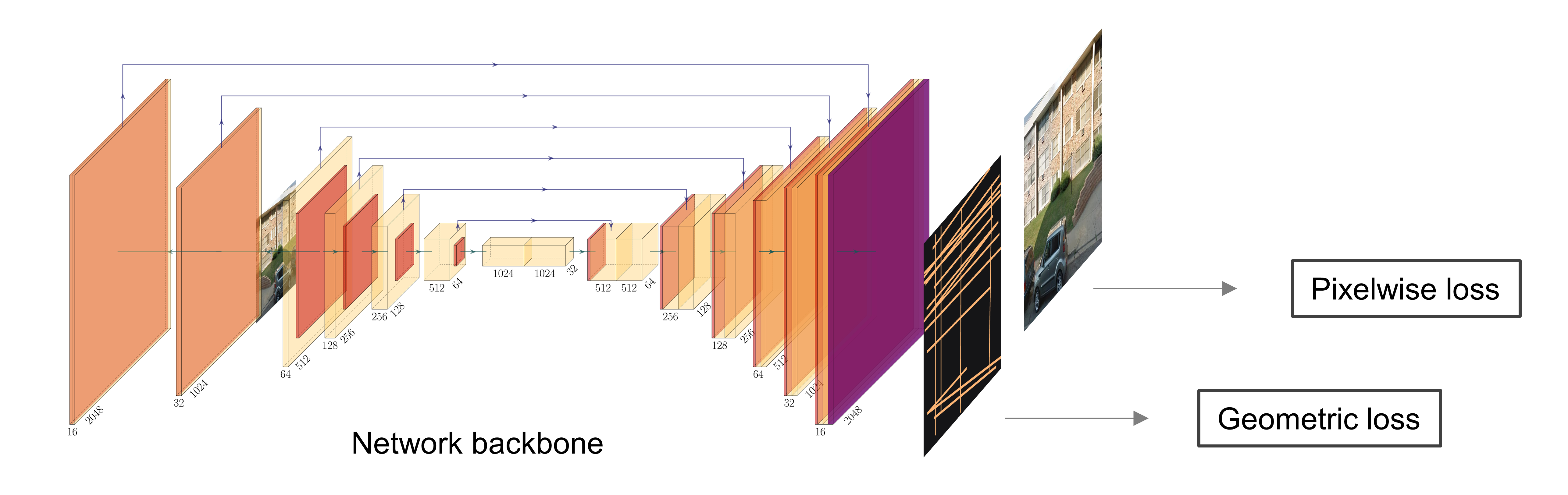}
	\caption{Compared with previous methods, we introduce a network that constrains geometric loss and pixel loss, effectively preserving the image structure.}
	\label{fig:arc} 
\end{figure}

This paper presents a deep-learning method for city scene super-resolution reconstruction, focusing on preserving the geometric features of super-resolution images by extracting key point and line features from regular geometric objects in urban scenes.
As shown in Figure. \ref{fig:arc}, the proposed method comprises three essential parts: the network structure for super-resolution modelling, geometric feature extraction for straight lines and regular shape detection, and geometric align loss for quantifying geometric errors in super-resolution images. The geometric-aware method aims to enhance the presentation of cultural heritage within city scenes.

\subsection{Network Structure}

\begin{figure}[htbp]
	\centering
	\includegraphics[width=\linewidth]{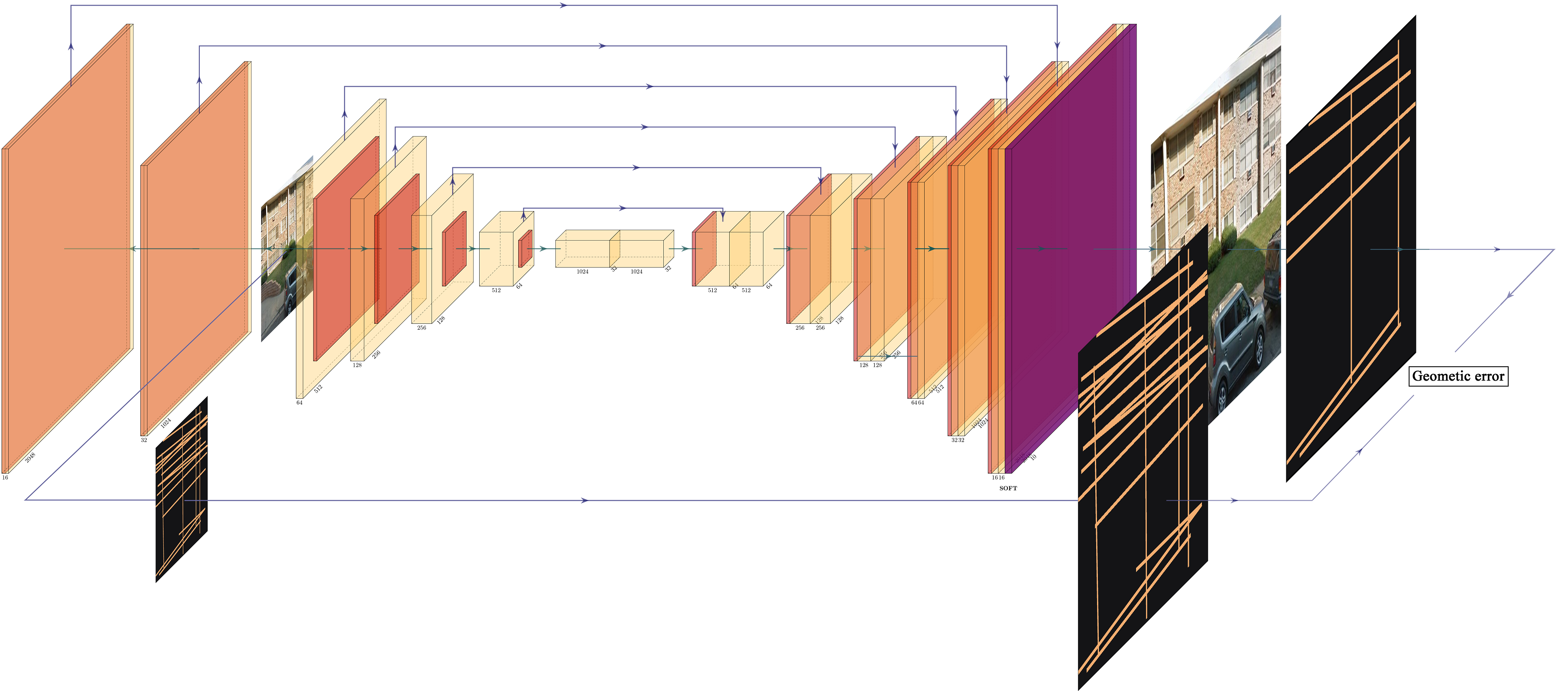}
	\caption{The framework of the GeoSR model derives from the UnetSR~\cite{lu2022single}. The notable modification is to align geometric features between low-resolution and high-resolution images.}
	\label{fig:architecture} 
\end{figure}

The proposed super-resolution method employs a modified UnetSR \cite{lu2022single}, a widely spread U-net architecture for super-resolution tasks, incorporating geometric feature constraints at low- and high-resolution levels.

As shown in Figure \ref{fig:architecture}, the proposed method has two main modules: 1) UnetSR-based super-resolution model and 2) geometric alignment constraint.
Similar to the U-net architecture\cite{ronneberger2015u}, UnetSR employs a contracting path on the left side for feature extraction. This path involves a 3$\times$3 kernel convolution, followed by rectified linear unit (ReLU) layers, and concludes with 2$\times$2 max-pooling operations with a stride of 2 for down-sampling. On the right side of Figure  \ref{fig:architecture}, the expanding path facilitates decoding. Each block in the expansive path includes up-sampling of the feature map, followed by a 2$\times$2 kernel that reduces the number of feature maps by half, and a subsequent 3$\times$3 convolution kernel followed by a ReLU layer. Compared with the original U-Net \cite{ronneberger2015u}, the UnetSR adds extra up-scaling layers in the feature extraction module, which aligns with the decoding module depth. Furthermore, the input image is up-scaled to larger sizes, establishing a high-level convolution layer on a larger scale. This block features a skip connection with the output block at the same network depth. The direct up-scaling of images mitigates errors arising from redundant complexity, resulting in outputs that closely align with ground truth. Each addition of an up-scale layer corresponds to a doubling in size. In practical terms, the network should incorporate n up-scale layers when the up-scale size is $2^{n}$.

Regarding the calculation of geometric alignment error, the proposed method extracts geometric information from the original image, projects it onto a large-scale image, and calculates the discrepancy between the geometric features of the super-resolution and the projection images.
The proposed method combines pixel-level relationships and geometric information to achieve high-quality super-resolution results.

\subsection{Geometric Feature Extraction}

GeoSR adopts a combination of Canny edge detection~\cite{canny1986computational} and Hough Transform~\cite{duda1972use} for geometric feature extraction in city scenes, owing to the rich regular geometric shapes and linear structures.

\begin{figure}[htbp]
	\centering
	\includegraphics[width=\linewidth]{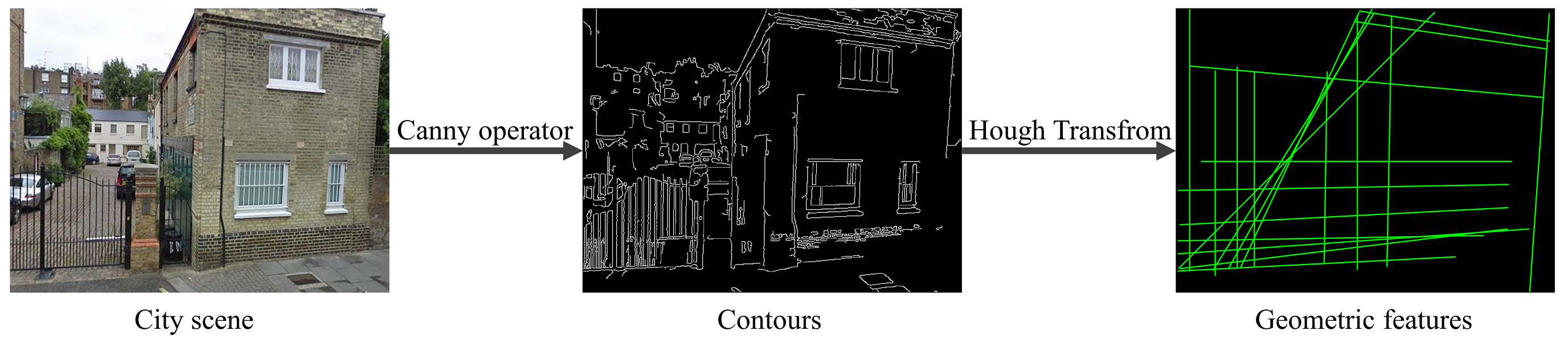}
	\caption{Diagram of the process of geometric feature extraction, involving canny edge convolution and the Hough transform.}
	\label{fig:geoextrac}
\end{figure}

Geometric feature processing, as illustrated in Figure \ref{fig:geoextrac} through the use of the Canny operator and the Hough transform, is critical in city scene image analysis. 
The Canny operator, an efficient edge detection algorithm, helps to isolate structural information by enhancing the clarity of boundaries and contours in images. 
Next, the Hough transform is applied to extracted contours, proficient in capturing regular geometric structures like straight lines and more complex shapes. 
The two-step process provides a structure-aware representation of urban images, facilitating the recognition of complex cityscapes. 
Given the intricate nature of cityscapes, which involve rigorous geometric constructs, the proposed method ensures a robust approach for analyzing structures within urban images.

First, Canny edge detection involves the intensity gradients computation with a Sobel operator. The gradient magnitude $G$ at a given pixel is formulate as:

\begin{equation}
	G(i,j)=\sqrt{G_{x}^{2}(i,j)+G_{y}^{2}(i,j)}
\end{equation}
where $G_x$ and $G_y$ are the gradients in the $x$ and $y$ directions, calculated by convolving the image with Sobel kernels\cite{kanopoulos1988design}. 
The gradient map $G$ in $x$ and $y$ direction of the ground truth image $Y$ shows below:

\begin{align}
	G_{x}&=Y*S_{x}\\
	G_{y}&=Y*S_{y}=Y*S_{x}^\top
\end{align}
where $*$ is the convolution operation, and $S_{x}$ donates the Sobel kernel in $x$ direction, and shown as:

\begin{equation}
	S_x=\begin{bmatrix}
		-1 & -2 & -1\\
		0 & 0 & 0\\
		1 & 2 & 1
	\end{bmatrix}
\end{equation}

Following edge detection, the Hough Transform is applied to recognize straight lines that frequently appear in typical cityscapes.
First, initialize an accumulator array $ H(\theta, \rho) $ to zero, where $\theta$ ranges from 0 to 180 degrees and $\rho$ ranges from $-D$ to $D$, where $D$ is the diagonal length of the image.
This parameterization allows the representation of lines in polar coordinates.
Next, for each non-zero pixel $(x, y)$ (edge pixel) in the image, we perform the following computations for each value of $\theta$ (incremented in small steps, e.g., 1 degree):

\begin{equation}
\rho = x  \cos(\theta) + y  \sin(\theta)
\end{equation}

Then, increment the accumulator $H(\theta, \rho)$ by one until the traversal is complete.
For the peak detection step, we identify the bins in the accumulator array with the highest values. These bins correspond to the parameters of the most prominent lines in the image. Therefore, we set a threshold to identify peaks in the Hough space.
For each peak $( \theta, \rho )$ in the Hough space, convert it back to the Cartesian coordinate system to obtain the line parameters and draw the lines on the image. The equations to obtain the line parameters are:

\begin{align}
x &= \rho  \cos(\theta) - t  \sin(\theta)\\
y &= \rho  \sin(\theta) + t  \cos(\theta)
\end{align}
where $t$ is a parameter that ranges over all real numbers.

Despite the inherent limitations of low-resolution input and the architectural diversity of city scenes, Canny edge detection and Hough Transform techniques combine to facilitate geometric feature extraction. 
The operation is an essential pre-processing step for super-resolution images while preserving critical geometric features, resulting in high-quality outputs.

\subsection{Geometric constraint}

The geometric loss function has two components: 1) the geometric classic error between super-resolution images and ground-truth images, and 2) the geometric align error between super-resolution images and the projections from low-resolution images.

\subsubsection{Classic geometric error}

The classic geometric error is defined as the error between the super-resolution image $I_{SR}$ and the ground-truth high-resolution image $I_{HR}$. The geometric features are exacted from pixel-wise images by Hough transform. Formally, the classic geometric loss $\mathcal{L}_{c}$ can be formulated as:

\begin{equation}
\mathcal{L}_{c} = || G_{SR} - G_{HR} ||_2
\end{equation}
where $G_{SR}$ donates the geometric map of super-resolution image, and $G_{HR}$ donates the geometric map of ground-truth high-resolution image.

\subsubsection{Geometric align error}

\begin{figure}[htbp]
	\centering
	\includegraphics[width=\linewidth]{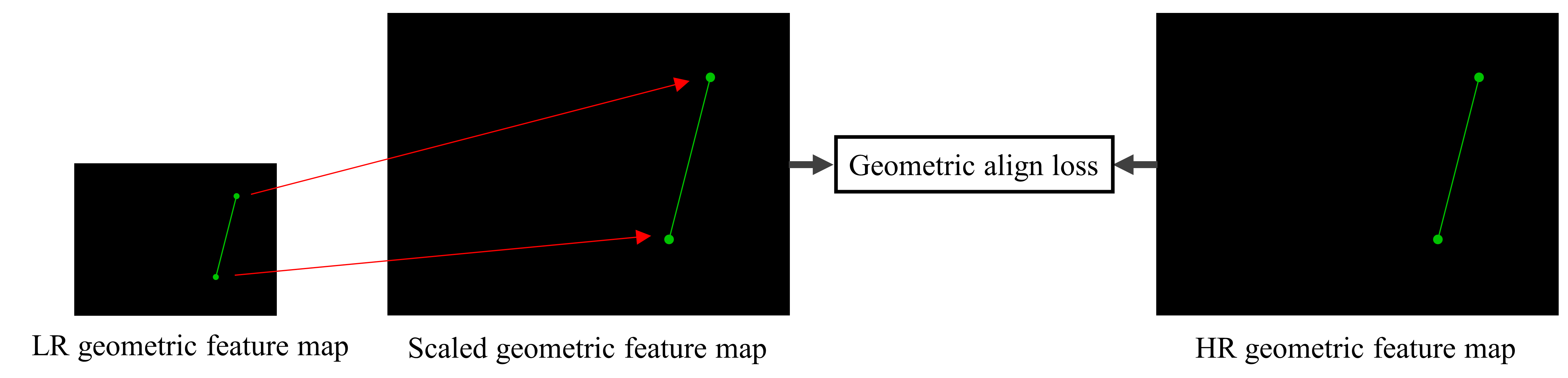}
	\caption{The align processing with low-resolution and high-resolution geometric feature map.}
	\label{fig:projprocessing}
\end{figure}

In order to preserve the geometric consistency between the low-resolution input, high-resolution output, and the ground-truth high-resolution image, this work proposes a structure-level loss function, namely geometric align loss $\mathcal{L}_{a}$. 
As shown in Figure \ref{fig:projprocessing}, the geometry-aware loss function compares the geometric features of the super-resolution reconstructed image with the proportional projection of the low-resolution geometric features to the high-resolution image. 
In the geometric feature map up-scaling process, the first step is to map endpoints to the large-scale image.
Then, endpoints are connected with equal-width lines to reconstruct the scaled geometric feature map.

This geometric align error can be attributed to two sources: one is the geometric loss caused by the downscale operation, and the other is the geometric loss resulting from non-downscale operations, which is referred to as model loss.
The systematic error $\mathcal{L}_{d}$ resulting from the downscale operation can be represented as follows:

\begin{equation}
	\mathcal{L}_{d} = || G_{HR} - P(G_{LR}) ||_2
\end{equation}
where $G_{HR}$ donates the geometric map of ground-truth high-resolution image, $G_{LR}$ donates the geometric map of low-resolution image, and $P(\cdot)$ donates the proportional projection operation.

Therefore, the pure model loss $\mathcal{L}_{p}$ from non-downscale operations can be formulated as:

\begin{equation}
	\mathcal{L}_{p} = | \mathcal{L}_{c}- \mathcal{L}_{d} |
\end{equation}
where $| \cdot|$ represents the absolute operation to prevent the negative loss function.
The pure model loss aims to minimize the geometric error caused by CNN model, thereby preserving the geometric structures during the super-resolution process.

By combining geometric align error with the mean square error (MSE) loss, the hybrid function becomes:

\begin{equation}
	\mathcal{L} = \mathcal{L}_{MSE} + \lambda_d \mathcal{L}_{d} +\lambda_p \mathcal{L}_{p}
\end{equation}
where $\lambda$ is a hyper-parameter that balances the reconstruction loss and the geometric align loss.

The proposed method allows us to preserve intricate geometric details during the super-resolution process, leading to better preservation and presentation of architectural elements in city scenes. Our approach leverages the rich geometric and structural regularities inherent in urban scenes, presenting a promising avenue for city scene super-resolution.

\section{Experimental Results}

\subsection{Implementation details}

Existing SISR methods are evaluated on four datasets: SET14\cite{bevilacqua2012low}, BSD300\cite{martin2001database}, Cityscapes\cite{cordts2016cityscapes} and GSV-Cities\cite{ali2022gsv}. 
The SET14 dataset contains 14 common images, and BSD300 is a classical image dataset that contains 300 images ranging from natural images to object-specific ones such as plants, people and food.
Cityscapes is a large-scale database focusing on the semantic understanding of urban street scenes.
The Cityscapes dataset consists of around 5000 fine-annotated images and 20000 coarse-annotated ones.
The GSV-Cities datasets comprise about 530,000 street scene images from cities worldwide.
In dataset pre-processing, the ground-truth high-resolution images are down-scaled by bicubic interpolation~\cite{de1962bicubic} to generate low-resolution and high-resolution image pairs for training and testing. 
For the dataset division, we maintain the original train/test split as provided in the datasets.

In the training parameter set, the data batch is set to 1.
The proposed model is trained by Adam optimizer\cite{kingma2014adam} with $\beta_1 = 0.9$, $\beta_2 = 0.999$, $\epsilon= 10^{-8}$.
The learning rate is initially set to $10^{-3}$ and decreases to half every 30 epoch. For the training balance, the hyper-parameter $\lambda_d$ is set to $0.1$ and the $\lambda_p$ is set to $1$.
The PyTorch implements the models involved in comparisons on one RTX 3090Ti GPU.

\subsection{Ablation Experiments}

In ablation experiments, we investigated various thresholds in the Hough transform, which is the essential component in geometric feature extraction.
The Hough threshold parameter is a limiting factor in identifying peaks in the Hough transform matrix, thereby distinguishing significant lines within the image. 
The parameter filters out lower peaks below a specified value, mitigating the impact of noise and irrelevant features.
However, a high threshold may disregard relevant information, while a low one may incorporate excessive noise. 
It is crucial to hold the balance, as the threshold directly influences the accuracy of the urban scene super-resolution. 

\begin{figure}[htbp]
	\centering
	\includegraphics[width=\linewidth]{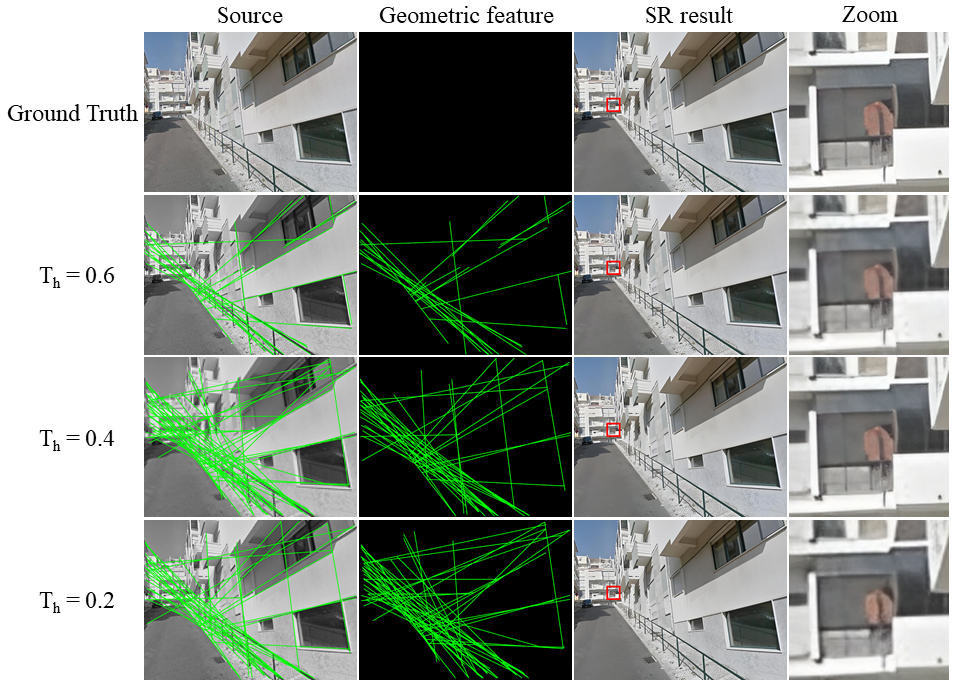}
	\caption{Visual results and geometric feature maps with various Hough transform thresholds.}
	\label{fig:ablation}
\end{figure}

\begin{table*}[htbp]
	\centering
	\caption{Ablation experiments on the GSV-Cities dataset ($\times$ 2)}
	{\begin{tabular}{llll}
			\hline
			\multirow{2}{*}{Network} & \multirow{2}{*}{$T_h$} & \multicolumn{2}{c}{GSV-Cities} \\
			&                        & PSNR           & SSIM          \\ \hline
			ResNet18                 & -                      & 34.8054        & 0.9602        \\
			ResNet18                 & 0.2                    & 36.9485        & 0.9705        \\
			ResNet18                 & 0.4                    & 38.7601        & 0.9701        \\
			ResNet18                 & 0.6                    & 36.9332        & 0.9689        \\ \hline
			UnetSR                   & -                      & 36.6461        & 0.9681        \\
			UnetSR                   & 0.2                    & 38.8955        & 0.9718        \\
			UnetSR                   & 0.4                    & 40.7950        & 0.9833        \\
			UnetSR                   & 0.6                    & 38.9512        & 0.9769        \\ \hline
	\end{tabular}}
	\label{tab4ablation}
\end{table*}

As shown in Table \ref{tab4ablation}, we observe the influence of distinct Hough threshold ($T_h$) values on two networks, ResNet18\cite{he2016deep} and UnetSR\cite{lu2022single}, across the GSV-Cities dataset.
Without geometric constraints, the baseline performances are optimized by mean square error. 
As the $T_h$ is incremented from 0.2 to 0.6, a notable trend emerged: an intermediate threshold value (0.4 in this case) enhanced the PSNR and SSIM metrics, while low or high values potentially introduced noise or overlooked essential information, thereby reducing performance. 
Thus, proper threshold selection is vital for optimal geometric feature extraction and superior super-resolution results in urban scenarios.

As depicted in Figure \ref{fig:ablation}, visual ablation results are presented corresponding to various Hough transform thresholds.
Observations show that the distinct improvement in image clarity is noticeable at a $T_h$ of 0.4. 
The coat's contour on the balcony becomes defined, and the corresponding geometric feature map is precise. 
In contrast, the geometric feature map with a $T_h$ of 0.2 appears chaotic. 
These qualitative observations prove that proper threshold selection produces accurate super-resolution results, especially in urban scenes with geometric regularities.

\subsection{Comparison with state-of-the-art Results}

To assess the model quality, we evaluated the proposed method against existing methods, comprising {\bf SRCNN}~\cite{dong2015image}, {\bf FSRCNN}~\cite{dong2016accelerating}, {\bf VDSR}~\cite{kim2016accurate}, {\bf CARN}~\cite{ahn2018fast}, {\bf DRCN}~\cite{kim2016deeply}, {\bf SRGAN}~\cite{ledig2017photo}, {\bf ESPCN}~\cite{shi2016real}, {\bf EDSR}~\cite{lim2017enhanced}, {\bf FKP}~\cite{liang2021flow}, {\bf NLSAN}~\cite{mei2021image} and {\bf DBPN}~\cite{haris2018deep}, on SET14\cite{bevilacqua2012low}, BSD300\cite{martin2001database}, Cityscapes\cite{cordts2016cityscapes}, and GSV-Cities\cite{ali2022gsv} datasets.
For fair comparisons, we re-implement existing networks with author's project and default model parameters.
Due to the missing implement of $\times$8 enlargement on some neural networks, namely {\bf SRCNN}~\cite{dong2015image}, {\bf FSRCNN}~\cite{dong2016accelerating}, {\bf VDSR}~\cite{kim2016accurate}, {\bf EDSR}~\cite{lim2017enhanced}, CARN~\cite{ahn2018fast}, FKP~\cite{liang2021flow}, and NLSAN~\cite{mei2021image}, we re-train such models for the super-resolution tasks at high magnification.
In quantitative comparisons, the {\bf bicubic}~\cite{de1962bicubic} interpolation method works as the benchmark, to evaluate various deep-learning methods.

\begin{table}[htbp]
	\centering
	\caption{Comparison of accuracy, parameter number and running time($8\times$) on GSV-Cities dataset}
	{\begin{tabular}{lrrrrl}
			\hline
			Method		&	Params(M)& 	Time(ms)	&	PSNR 		& 	SSIM	\\
			\hline
			Bicubic~\cite{de1962bicubic}	&	 -		&	- 			&	20.3965	&	0.6403	\\
			ESPCN~\cite{shi2016real}		&	 0.08 	&	1.2980		&	22.0324	&	0.7136	\\
			SRCNN~\cite{dong2015image}		&	 0.17	&	3.0606		&	20.2303	&	0.6846\\
			VDSR~\cite{kim2016accurate}		&	 0.22	&	6.9123		&	22.0039	&	0.7141\\
			EDSR~\cite{lim2017enhanced}		&	 0.78	&	16.5483		&	23.4584	&	0.7723\\
			FSRCNN~\cite{dong2016accelerating}	&	 0.03	&	0.8891		&	21.2715	&	0.6629	\\
			DRCN~\cite{kim2016deeply}		&	 0.11	&	60.5737		&	23.4698 &	0.7727\\
			SRGAN~\cite{ledig2017photo}		&	 6.54	&	12.0300		&	22.4210	&	0.7210	\\
			DBPN~\cite{haris2018deep}		&	 23.21	&	82.5170		&	23.8859 &	0.7867\\
			CARN~\cite{ahn2018fast}			&	 1.72	&	3.2510		&	23.5236	&	0.7786\\
			FKP	~\cite{liang2021flow}		&	 0.15	&	7.8566		&	23.0153	&	0.7705\\
			NLSAN~\cite{mei2021image}		&	 10.12	&	26.5083		&	23.9134	&	0.7845\\
			UnetSR\cite{lu2022single}		&	 8.50	&	16.9530		&{\color{blue}24.8691}	&{\color{blue}0.7974}	\\
			GeoSR							&	 8.50	&	16.9708		&{\color{red}26.0288} &	{\color{red}0.8201}\\
			\hline
			\multicolumn{5}{p{220pt}}{Red numbers mark the best score.}\\
			\multicolumn{5}{p{220pt}}{Blue numbers mark the second best score.}\\
			\multicolumn{5}{p{220pt}}{M is the abbreviation of Million.}\\
	\end{tabular}}
	\label{tab4sota}
\end{table}

Table.\ref{tab4sota} presents a comparative analysis of various super-resolution methods on the GSV-Cities dataset, focusing on parameters, runtime, and accuracy. GeoSR showcases superior performance with the highest PSNR of 26.0288 and SSIM of 0.8201, both marked in red. While UnetSR follows closely behind with the second-best scores (highlighted in blue) of 24.8691 for PSNR and 0.7974 for SSIM, many other methods, such as ESPCN, SRCNN, and VDSR, lag in performance metrics. Furthermore, while methods like DBPN have significantly more parameters (23.21M), they do not necessarily translate to the best results, emphasizing GeoSR's effectiveness.

\begin{table*}[htbp]
	\centering
	\caption{Accuracy comparison on SET14, BSD300, Cityscapes and GSV-Cities dataset}
	\begin{tabular}{lcccccllll}
		\hline
		\multirow{2}{*}{Method}                                  & \multirow{2}{*}{Scale} & \multicolumn{2}{c}{SET14}                                                          & \multicolumn{2}{c}{BSD300}                                                           & \multicolumn{2}{c}{Cityscapes}                      & \multicolumn{2}{c}{GSV-Cities}                      \\
		&                        & PSNR                                     & SSIM                                    & PSNR                                      & SSIM                                     & \multicolumn{1}{c}{PSNR} & \multicolumn{1}{c}{SSIM} & \multicolumn{1}{c}{PSNR} & \multicolumn{1}{c}{SSIM} \\ \hline
		Bicubic\cite{de1962bicubic}       & $\times$2              & 24.4523                                  & 0.8482                                  & 26.6538                                   & 0.7924                                   & 31.2861                  & 0.8893                   & 29.2643                  & 0.9199                   \\
		ESPCN\cite{shi2016real}           & $\times$2              & 26.7606                                  & 0.8999                                  & 28.9832                                   & 0.8732                                   & 33.8239                  & 0.9104                   & 34.1060                  & 0.9511                   \\
		SRCNN\cite{dong2015image}         & $\times$2              & 25.9711                                  & 0.8681                                  & 28.6943                                   & 0.8671                                   & 33.5075                  & 0.9095                   & 33.8277                  & 0.9518                   \\
		VDSR\cite{kim2016accurate}        & $\times$2              & 28.6617                                  & {\color{red} 0.9269} & 29.3889                                   & 0.8785                                   & 34.4207                  & 0.9234                   & 35.3600                  & 0.9604                   \\
		EDSR\cite{lim2017enhanced}        & $\times$2              & 24.0624                                  & 0.8383                                  & 28.3119                                   & 0.8621                                   & 32.7795                  & 0.9119                   & 37.1469                  & 0.9626                   \\
		FSRCNN\cite{dong2016accelerating} & $\times$2              & 23.1284                                  & 0.8123                                  & 28.7534                                   & 0.8681                                   & 33.3006                  & 0.9215                   & 34.8375                  & 0.9564                   \\
		DRCN\cite{kim2016deeply}          & $\times$2              & 24.4234                                  & 0.8458                                  & 27.5089                                   & 0.8088                                   & 32.0957                  & 0.9047                   & 36.2425                  & 0.9575                   \\
		SRGAN\cite{ledig2017photo}        & $\times$2              & 23.9553                                  & 0.8195                                  & 28.7072                                   & 0.8633                                   & 31.6192                  & 0.8998                   & 35.6830                  & 0.9550                   \\
		DBPN\cite{haris2018deep}          & $\times$2              & 28.4092                                  & 0.9196                                  & 29.8675                                   & {\color{red}0.8834}   & 34.4227                  & 0.9260                   & 36.9533                  & 0.9628                   \\
		CARN\cite{ahn2018fast}            & $\times$2              & {\color{blue}29.1020} & 0.9135                                  & {\color{blue} 30.1520} & 0.8789                                   & 34.5918                  & 0.9248                   & 36.8421                  & 0.9598                   \\
		FKP\cite{liang2021flow}           & $\times$2              & 28.1251                                  & 0.8868                                  & 28.6510                                   & 0.8720                                   & 33.2110                  & 0.9169                   & 35.7520                  & 0.9611                   \\
		NLSAN\cite{mei2021image}          & $\times$2              & {\color{red}29.1250}  & 0.9177                                  & {\color{red} 30.1730}  & 0.8812                                   & 34.1563                  & 0.9186                   & 37.4604                  & 0.9645                   \\
		UnetSR\cite{lu2022single}                                                  & $\times$2              & 28.3965                                  & {\color{blue}0.9198} & 29.8403                                   & {\color{blue} 0.8816} & {\color{blue}35.4989}                  & {\color{blue}0.9530}                   & {\color{blue}38.0494}                  & {\color{blue}0.9651}                   \\
		GeoSR                                                    & $\times$2              & 28.8460                                  & 0.9176                                  & 29.9411                                   & 0.8763                                   & {\color{red}37.5564}                  & {\color{red}0.9557}                   & {\color{red}40.7950}        & {\color{red}0.9833}                   \\ \hline
		Bicubic\cite{de1962bicubic}       & $\times$4              & 19.7167                                  & 0.6089                                  & 23.5053                                   & 0.6157                                   & 26.7078                  & 0.7757                   & 23.6503                  & 0.7748                   \\
		ESPCN\cite{shi2016real}           & $\times$4              & 20.6292                                  & 0.6333                                  & 24.4899                                   & 0.6641                                   & 28.0118                  & 0.8091                   & 26.5639                  & 0.8472                   \\
		SRCNN\cite{dong2015image}         & $\times$4              & 20.5825                                  & 0.6288                                  & 24.2232                                   & 0.6597                                   & 26.7811                  & 0.7546                   & 26.2039                  & 0.8427                   \\
		VDSR\cite{kim2016accurate}        & $\times$4              & 21.4763                                  & 0.6991                                  & 24.7077                                   & 0.6816                                   & 29.0004                  & 0.8196                   & 27.3933                  & 0.8737                   \\
		EDSR\cite{lim2017enhanced}        & $\times$4              & 19.9784                                  & 0.6269                                  & 23.9192                                   & 0.6513                                   & 26.5737                  & 0.6994                   & 29.4719                  & 0.9008                   \\
		FSRCNN\cite{dong2016accelerating} & $\times$4              & 19.3255                                  & 0.5941                                  & 24.2499                                   & 0.6599                                   & 26.6219                  & 0.7537                   & 26.5053                  & 0.8481                   \\
		DRCN\cite{kim2016deeply}          & $\times$4              & 19.7077                                  & 0.6078                                  & 23.3462                                   & 0.6132                                   & 26.3315                  & 0.7648                   & 28.6425                  & 0.8875                   \\
		SRGAN\cite{ledig2017photo}        & $\times$4              & 19.3877                                  & 0.5976                                  & 24.1675                                   & 0.6485                                   & 26.1825                  & 0.7539                   & 26.9069                  & 0.8688                   \\
		DBPN\cite{haris2018deep}          & $\times$4              & {\color{red}21.7657}  & {\color{red}0.7171}  & 25.0644                                   & {\color{blue}0.6967}  & 28.3890                  & 0.8101                   & 29.3664                  & 0.9000                   \\
		CARN\cite{ahn2018fast}            & $\times$4              & 20.5241                                  & 0.7054                                  & {\color{blue}26.2536}  & 0.6825                                   & 28.7415                  & 0.8177                   & 28.3351                  & 0.8841                   \\
		FKP\cite{liang2021flow}           & $\times$4              & 21.6523                                  & 0.7082                                  & 26.1022                                   & 0.6900                                   & 28.9646                  & 0.8075                   & 26.5131                  & 0.8416                   \\
		NLSAN\cite{mei2021image}          & $\times$4              & 21.6805                                  & 0.7103                                  & {\color{red}26.2933}   & {\color{red}0.6983}   & 29.1033                  & 0.8234                   & 29.6497                  & 0.9034                   \\
		UnetSR\cite{lu2022single}                                                 & $\times$4              & {\color{blue}21.6825} & {\color{blue}0.7112} & 24.9522                                   & 0.6901                                   & {\color{blue}30.3018}                  & {\color{blue}0.8765}                   & {\color{blue}30.0970}                  & {\color{blue}0.9063}                   \\
		GeoSR                                                   & $\times$4              & 21.3069                                  & 0.6978                                 & 26.0210                                   & 0.6896                                   & {\color{red}32.0578}                 & {\color{red}0.9059}                   & {\color{red}34.6824}              & {\color{red}0.9504}                   \\ \hline
		Bicubic\cite{de1962bicubic}       & $\times$8              & 16.1132                                  & 0.3673                                  & 21.3115                                   & 0.4933                                   & 23.1663                  & 0.6729                   & 20.3965                  & 0.6403                   \\
		ESPCN\cite{shi2016real}           & $\times$8              & 16.3441                                  & 0.3628                                  & 21.6447                                   & 0.5064                                   & 23.8575                  & 0.6860                   & 22.0324                  & 0.7136                   \\
		SRCNN\cite{dong2015image}         & $\times$8              & 16.3853                                  & 0.3614                                  & 21.8101                                   & 0.5075                                   & 21.4967                  & 0.6011                   & 20.2303                  & 0.6846                   \\
		VDSR\cite{kim2016accurate}        & $\times$8              & 16.7994                                  & 0.4095                                  & 21.9697                                   & 0.5181                                   & 24.3488                  & 0.6997                   & 22.0039                  & 0.7141                   \\
		EDSR\cite{lim2017enhanced}        & $\times$8              & 15.7257                                  & 0.3209                                  & 21.6573                                   & 0.5067                                   & 22.3799                  & 0.5897                   & 23.4584                  & 0.7723                   \\
		FSRCNN\cite{dong2016accelerating} & $\times$8              & 14.5788                                  & 0.2541                                  & 21.3311                                   & 0.5011                                   & 21.4435                  & 0.6063                   & 21.2715                  & 0.6629                   \\
		DRCN\cite{kim2016deeply}          & $\times$8              & 16.1497                                  & 0.3685                                  & 21.2771                                   & 0.4934                                   & 23.0433                  & 0.6624                   & 23.4698                  & 0.7727                   \\
		SRGAN\cite{ledig2017photo}        & $\times$8              & 15.7133                                  & 0.3221                                  & 21.8766                                   & 0.5121                                   & 22.3840                  & 0.6329                   & 22.4210                  & 0.7210                   \\
		DBPN\cite{haris2018deep}          & $\times$8              & 16.7398                                  & {\color{red}0.4122}  & 22.0577                                   & 0.5229  & 25.0308                  & 0.7088                   & 23.8859                  & 0.7867                   \\
		CARN\cite{ahn2018fast}            & $\times$8              & 17.7012                                  & 0.4058                                  & {\color{blue}22.6811}   & 0.4913                                   & 24.8379                  & 0.6808                   & 23.5236                  & 0.7786                   \\
		FKP\cite{liang2021flow}           & $\times$8              & 17.0810                                  & 0.4010                                  & 21.9560                                   & 0.5134                                   & 26.5032                  & 0.7684                   & 23.0153                  & 0.7705                   \\
		NLSAN\cite{mei2021image}          & $\times$8              & {\color{red}17.8780}  & 0.4092                                  & 21.9548                                   & 0.5094                                   & 26.4946                  & 0.7663                   & 23.9134                  & 0.7845                   \\
		UnetSR\cite{lu2022single}               & $\times$8              & {\color{blue}17.8289} & {\color{blue}0.4103} & 22.0368  & {\color{blue}0.5235}   & {\color{blue}26.8386}                  & {\color{blue}0.7979}                   & {\color{blue}24.8691}                  & {\color{blue}0.7974}                   \\
		GeoSR                                                    & $\times$8              & 17.1207                                  & 0.4008                                  & {\color{red}22.1530}                                   & {\color{red}0.5296}                                   & {\color{red}27.0530}                  & {\color{red}0.8132}                   & {\color{red}26.0288}                  & {\color{red}0.8201}                   \\ \hline
	\end{tabular}
	\label{tab4main}
\end{table*}

Table.\ref{tab4main} demonstrates the accuracy comparison of the existing models on SET14, BSD300, Cityscapes and GSV-Cities datasets.
In Table.\ref{tab4main}, GeoSR outperforms various super-resolution methods across various scales. 
The results illustrate GeoSR's superior performance with regard to PSNR and SSIM, especially notable on larger scales ($\times$4 and $\times$8) and complex city scene datasets such as Cityscapes and GSV-Cities.
In particular, GeoSR attains the highest PSNR and SSIM scores on city-specific datasets at $\times$2, $\times$4, and $\times$8 scales, illustrating the model's significant capability in image super-resolution.
The effectiveness of GeoSR can be attributed to geometric constraints, which help to extract structural information, especially in complex urban scenes. 
For instance, the superiority of GeoSR is more apparent in Cityscapes and GSV-Cities datasets, which contain numerous regular geometric shapes. 
The performance of the GeoSR on the SET14 and BSD300 datasets, while not the leading result, still indicates considerable effectiveness. 
Nevertheless, the proposed method's ability to deliver high accuracy in diverse conditions proves its versatility.

\begin{figure}[htbp]
	\centering
	\includegraphics[width=\linewidth]{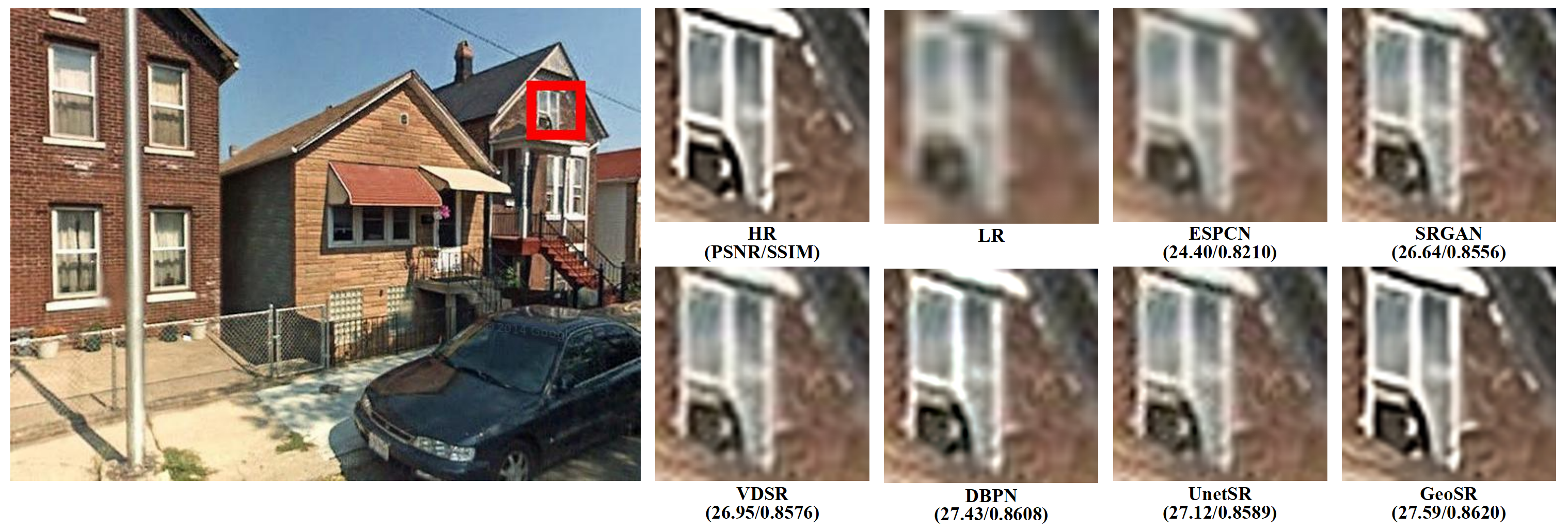}
	\caption{Super-resolution results on GSV-Cities dataset ($\times 2$).}
	\label{fig:demo1}
\end{figure}

\begin{figure}[htbp]
	\centering
	\includegraphics[width=\linewidth]{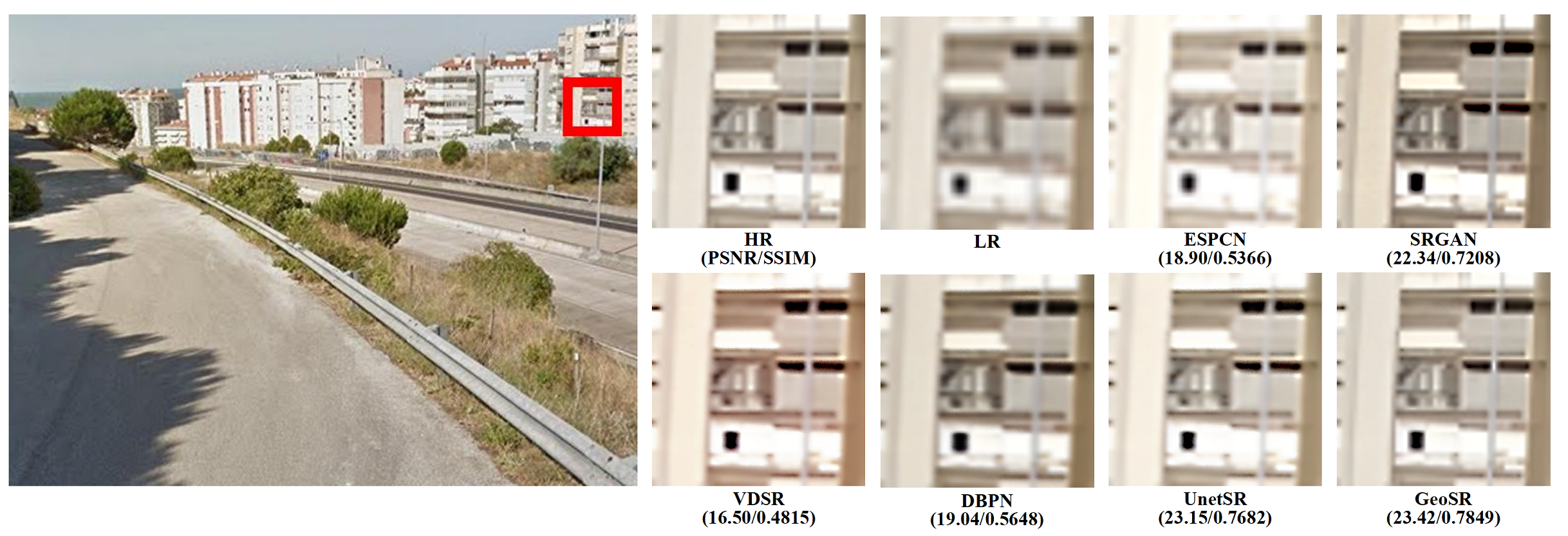}
	\caption{Super-resolution results on GSV-Cities dataset ($\times 4$).}
	\label{fig:demo2}
\end{figure}

\begin{figure}[htbp]
	\centering
	\includegraphics[width=\linewidth]{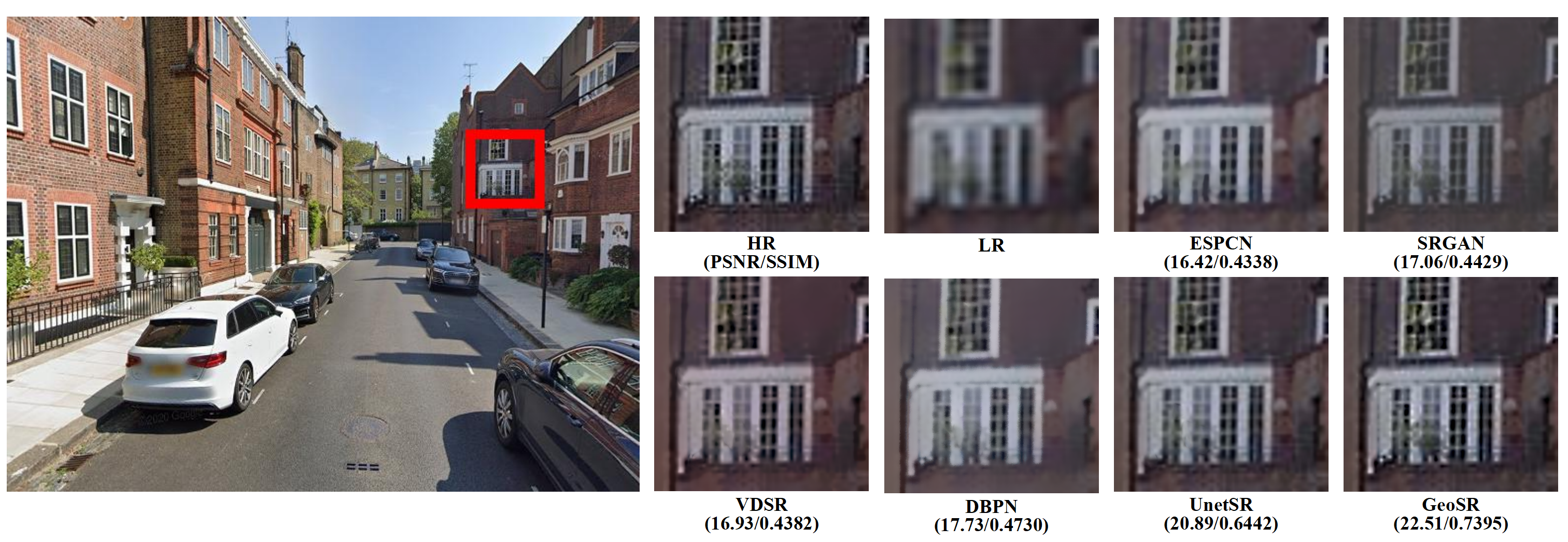}
	\caption{Super-resolution results on GSV-Cities dataset ($\times 8$).}
	\label{fig:demo3}
\end{figure}

The qualitative results, as illustrated in Figures \ref{fig:demo1} , \ref{fig:demo2}, and \ref{fig:demo3}, confirm GeoSR's superiority across all scales. As depicted in Figure ~\ref{fig:demo1}, the window edges are sharper than other reconstruction results. In Figure ~\ref{fig:demo2}, the boundaries of the shadow on the balcony are well-defined, with obvious gradients. The street scene shown in Figure ~\ref{fig:demo1} demonstrates GeoSR's ability to reconstruct complex details, such as the greenery on the window sill, preserving natural appearances. Furthermore, the framework of the upper window appears as regular as the ground truth.

In summary, the precise reconstruction across various scenes showcases that GeoSR is efficient in super-resolution, particularly in structurally complex urban scenes. 
The model's enhancement stems from the incorporation of dominant lines extraction. The Hough Transform complements the SR process by preserving significant structural elements, thereby contributing to a more coherent and visually faithful reconstruction.
The constraint on dominant edges is justified as it helps maintain the global structural integrity of the scene, impacting not only fine details but also ensuring a more contextually accurate representation.

\section{Conclusion}

This paper proposes a novel single-image super-resolution method, explicitly tailored for urban scenes, focusing on preserving geometric structures and enhancing the cultural heritage representation. Using Hough Transform, we extract regular geometric objects and lines abundant in cityscapes, enabling the computation of geometric errors between low-resolution and high-resolution images. The proposed method minimizes geometric errors during the super-resolution process, resulting in higher-resolution images that maintain essential geometric regularities while enhancing pixel-level details. Extensive validations on the Cityscapes and GSV-Cities datasets demonstrate the superiority of our approach over existing state-of-the-art methods, particularly in complex city scenes where geometric preservation is crucial for accurate representation.

In future work, we will explore additional geometric constraints and regularization techniques to enhance our method's accuracy and robustness. Integration of domain-specific knowledge and scene-specific priors will adapt the super-resolution process for different urban environments. Incorporating self-supervised techniques may improve generalization to handle diverse city scenes. Extending the application of our approach to cultural heritage preservation, urban planning, and environmental monitoring will provide valuable insights and practical solutions for real-world applications.

\subsection* {Code, Data, and Materials Availability} 
The project is openly available at \url{https://github.com/Mnster00/GeoSR}.

\subsection* {Acknowledgments}
This work is supported by National Social Science Fund of China Major Project in Artistic Studies (No.22ZD18), and China Postdoctoral Science Foundation (No.2023M741411).


\bibliographystyle{spiejour}   




\listoffigures
\listoftables

\end{spacing}
\end{document}